# World Models: The Safety Perspective

Zifan Zeng[1,2], Chongzhe Zhang[1,3], Feng Liu[1], Joseph Sifakis[4], Qunli Zhang[1], Shiming Liu[5] and Peng Wang[5]

*Abstract*—With the proliferation of the Large Language Model (LLM), the concept of World Models (WM) has recently attracted a great deal of attention in the AI research community, especially in the context of AI agents. It is arguably evolving into an essential foundation for building AI agent systems. A WM is intended to help the agent predict the future evolution of environmental states or help the agent fill in missing information so that it can plan its actions and behave safely. The safety property of WM plays a key role in their effective use in critical applications. In this work, we review and analyze the impacts of the current state-of-the-art in WM technology from the point of view of trustworthiness and safety based on a comprehensive survey and the fields of application envisaged. We provide an in-depth analysis of state-of-the-art WMs and derive technical research challenges and their impact in order to call on the research community to collaborate on improving the safety and trustworthiness of WM.

*Index Terms*—AI Safety, LLM, Embodied AI, World Model, Intelligent Agents

## I. INTRODUCTION

Recent years have witnessed rapid advancements in transformer-based generative models [1], with their capabilities expanding from natural language processing (NLP) to multimodal applications [2]. Frontier models such as SORA [3], LINGO-1 [4], and GAIA-1 [5] demonstrate an unprecedented ability to generate remarkably realistic videos, suggesting an initial grasp of fundamental worldly principles like physics and spatiotemporal continuity, achieved solely through training on video and language datasets. This emerging capability opens new avenues for research, as understanding world models is crucial for developing next-generation intelligent systems. The concept of data-driven world models, initially proposed in 2017 using recurrent neural network (RNN) or long-short-term memory (LSTM) architectures [6], showed promise but was limited by constraints in sequence length, memory, and parallel capability. These early experiments demonstrated only modest capabilities in relatively simple simulated gaming environments. The advent of transformer-based methods has led to significant improvements, with recent experimental results showing encouraging progress. Consequently, many contemporary AI agent architectures now incorporate world models as a cornerstone component [7]. Our paper focuses on world models for a specific class of agents known as embodied AI agents, which interact with the physical world. We examine these world models from the point of view of their safety, addressing a crucial gap in current research.

Auto-regressive generative models suffer from inherent deficiencies such as hallucination [8], [9]. This limitation poses significant risks in safety-critical applications like robotics and autonomous driving systems (ADS) [10], sparking controversial discussions [11]. Despite the current attention, we observe a lack of comprehensive analysis regarding the safety aspects of world models for embodied AI agents. Our paper aims to address this gap by providing a concise yet thorough review and investigation, followed by an in-depth analysis from a safety perspective. Finally, we identify high-priority research directions.

The main contributions of this paper are summarized as follows:

- We conduct a literature survey of recent achievements in world model research and show the development of techniques for realizing world models in chronological order.
- We address the identified safety issues of the world models on embodied AI applications such as autonomous driving.
- We propose possible approaches for future research to facilitate the future of trustworthy world models.

The paper is structured as follows: Section II presents a current definition of world models, offering an in-depth investigation and taxonomy of state-of-the-art approaches. We retrospectively review the development path of modern world model approaches across various application perspectives; Section III analyzes the safety-related deficiencies of current approaches from a critical perspective; Section IV proposes a research agenda highlighting high-priority topics to improve the safety of world models. By addressing these crucial aspects, we aim to bring clarity and contribute to the current debate on world models in embodied AI and promote the development of safer and more trustworthy intelligent systems.

## II. THE EVOLUTION OF WORLD MODELS

In this section, we present an analysis of the development of existing world models, with the intention of showing how the techniques used to implement world models have evolved over time. Fig. 1 gives an overview of the development in chronological order. To the best of our knowledge, the core techniques adopted in the world models can be divided into four categories, as shown in the figure. *Recurrent neural network (RNN)* is one of the initial ideas to implement works models. Due to the capability of processing sequential data, RNNs are widely used in reinforcement learning (RL) agents as an emulator to predict the transition of the world state in the next steps. *Transformer* is proposed initially for machine translation. However, the transformer-based models are found to perform better in dealing with longer contexts than RNN over various tasks. *Diffusion models* learn how to generate images from a standard Gaussian distribution. A backbone model such as U-Net [12] is trained to predict the noise $\epsilon_t$ given the noised image in the denoising process (reverse process). In addition, other techniques are used to build a world model that can predict the development of the world state. However, due to the limited number of such techniques, we will introduce them in the last subsection.

[1]Huawei Technologies Duesseldorf GmbH, RAMS Lab, Munich, Germany.
[2]Technical University of Munich, School of Computation, Information, and Technology, Garching, Germany. `zifan.zeng@tum.de`
[3]Technical University of Berlin, Faculty of Electrical Engineering and Computer Science, Berlin, Germany. `chongzhe.zhang@campus.tu-berlin.de`
[4]Univ. Grenoble Alpes, CNRS, Grenoble INP, VERIMAG and SUSTECH/RITAS, Grenoble, France.
[5]Huawei Technologies Co., Ltd., RAMS Lab, Shenzhen, China.

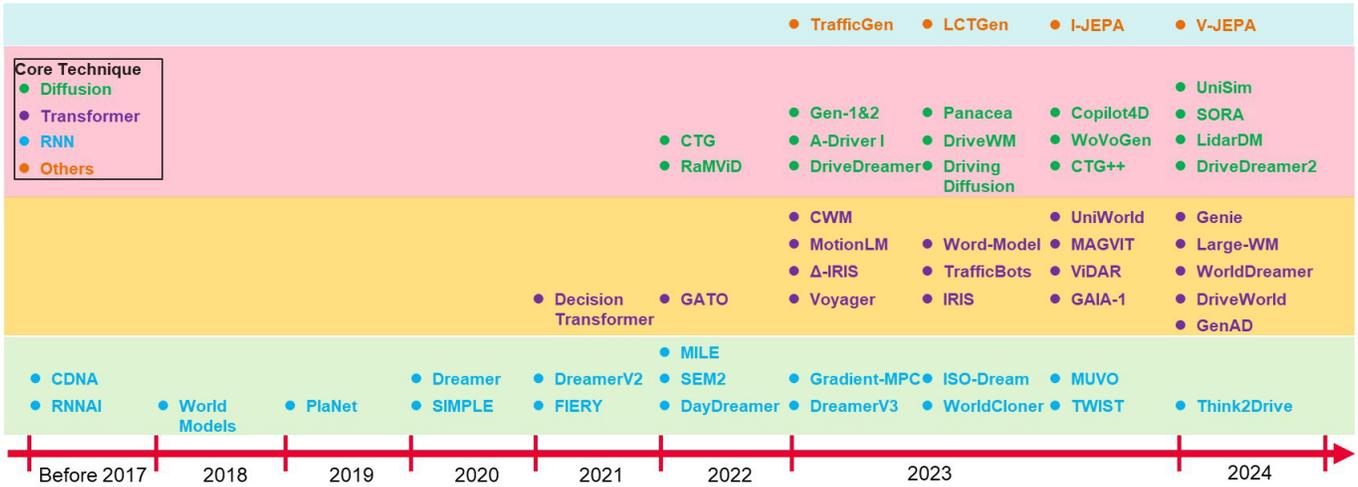

Fig. 1. The development of world models categorized with the core techniques in chronological order.

*A. RNN-Based Methods*

Schmidhuber et al. [13]–[16] have advocated in training world models using recurrent neural networks (RNN) to learn the temporal conversion of states since 1990. In 2018, Ha and Schmidhuber proposed an RL agent driven by a world model composed of a convolution neural network (CNN) and a recurrent neural network. In the proposed framework, the CNN encodes the image input from the interaction environment, and the RNN predicts the future states represented in the latent space.

The long-short-term memory (LSTM) [17] method further improved dealing with longer sequences than RNN. Kaiser et. al. introduced their Simulated Polich Learning (SimPLe) algorithm [18], where an LSTM model is deployed for future prediction of state embeddings.

In 2019, Hafner et al. [19] proposed the recurrent state space model (RSSM) architecture where both deterministic and stochastic variables are constructed to represent the states in latent space. RL agents driven by RSSM architecture achieved significant improvement of the performance in both interacting with virtual environments [20]–[22] like Atari Games [23], and learning how to complete basic embodied tasks in real-world [24] in 2022.

Hu et al. introduced the model-based imitation learning (MILE) approach for end-to-end autonomous driving in [25]. The MILE approach utilizes a network with both stochastic and deterministic variables for the representation of latent states, and the gated recurrent units are adopted to emulate the world state evolution. The model is trained in the imitation learning way, which tries to imitate the experts' behavior based on the collected observation-action pairs.

A huge pain point in training intelligent agents is the distribution shift between the training data and the real dynamic world. Most of the existing works suffer from poor generalization capability. Thus, their behavior degrades heavily in an unseen environment. WorldCloner was proposed by Balloch et al. in [26], trying to adapt to an open and dynamic world in a neuro-symbolic manner. A rule model that detected and updated the world rules was equipped in addition to a normal RL model. When the rules in the experimental NovGrid environment [27] change, the rule model will try to detect and learn the new world rules.

Besides the RNNs' major application in being equipped in RL or IL agents, Bogdoll and Yang et al. [28] tried to exploit the potential of RNN in generating new driving scenarios. Encoders were trained to encode initial frames into latent space, and gated recurrent units predicted the possible world states in the next steps based on the given driving actions. Multi-modal decoders not only decoded into high-resolution image data but also into point cloud and 3D space occupancy formats with 3D information.

*B. Transformer-Based Methods*

Decision Transformer [29] is a world model for playing Atari games. State, actions, and system feedback in the game are serialized as linear embedding, combined with a positional episodic timestep, and fed into the generative pre-trained transformer (GPT) architecture, generating new agent actions based on future desired returns via autoregression. GATO [30] is a multi-modal, multi-task, multi-embodiment generative world model. Multi-modal data from different tasks are serialized into flat sequences of tokens, and a transformer-based network is trained by completing the reconstruction of the masked inputs. In [31], [32], the GPT-like Transformer simulates the environment dynamics. It takes the state information and action predictions encoded by the autoencoder as inputs to make predictions about the next state, reward policy, and episode termination. The GPT-Like Transformer enables the game agent to be data-efficiently trained in the world model. Voyager [33] implements an LLM-driven lifelong learning agent. In Minecraft, the agent is trained through automated interactions with the LLM without access to model parameters and performs gradient-based training. The LLM with world knowledge automatically maximizes the search for learning curricula for the agent, creates executable behavioral code through an iterative prompting mechanism, and continuously enriches the skill base to improve the agent's ability to handle complex tasks.

Counterfactual World Model (CWM) [34] can be used as a unified model for visual computation of world models. Through structured mask recovery training, the model can grasp the key scene transition information through a few visual markers. Keypoints, optical flows, and object segments in computer vision can be constructed in CWM in the form of counterfactual prompts by examining the perturbed model response in a zero-shot fashion. In [35], a cognitive system covering probabilistic reasoning, logical and relational reasoning, visual and physical reasoning, as well as social reasoning about agents and their plans, is constructed by modeling thinking through LLM. The construction of a language-driven world model is realized.

In GenAD [36], BEV tokens are converted into agent tokens and map tokens through two different tasks. They are combined with additional ego tokens for self-attention and cross-attention, transforming the traffic scene into a specific instance-centric scene representation. This representation is then mapped into the trajectory space and decoded into the agent's future trajectory. In ViDAR [37], a new visual point cloud forecasting task is proposed that enables synergic learning of visual semantics, 3D structure, and temporal dynamics in a model. This task can be used as an access to generalized visual perception and prediction capabilities for world models. In MotionLM [38], the autopilot agents' motion trajectories are discretized into motion tokens, combined with scene embeddings in the autoregressive transformer decoder to obtain trajectory predictions for multiple agents. TrafficBots [39] goes a step further. It introduces navigation destinations that allow vehicles to interact with multi-intelligent agents with personalities generated from a world model, enabling fidelity simulation and prediction of self-driving vehicle motion. In GAIA-1 [5], a generative world model creates realistic driving scenarios with video, text, and action prompts. Controlling the finesse of ego vehicle behavior and scene features, the agents' behaviors are predicted by an autoregressive transformer network-based world model. In UniWorld [40], the world model is constructed by fine-tuning the 4D geometric occupancy model. This enables the estimation of missing information about the world state and the prediction of future states. A transformer is used to construct 2D to 3D perspective transformations that provide a unified feature representation for the world model.

MAGVIT [41] can generate videos generically and efficiently. It quantizes videos into spatial-temporal visual tokens by 3D tokenizer, creates embedding for multiple video generation tasks using different masks, and learns to predict target tokens by the bidirectional transformer. In [42], scalable learning using RingAttention and Blockwise transformer on a huge number of long videos and books makes it possible to shape a model's understanding of the physical world in situations that are not easily described in words. This creates the basis for world models with more reliable reasoning and broader understanding. Through unsupervised training with large video game datasets, Genie [43] can generate interactive virtual worlds based on multimodal descriptions. The spatial-temporal (ST) transformer implements the attention mechanism in space & time and becomes the potential action model in the world model. In DriveWorld [44], a vision task centered on 4D scene understanding is used for world model training. Temporal-aware latent dynamics and spatial-aware latent statics are learned through the Memory State-Space Model. DriveWorld improves the capability of world-knowledge-driven autonomous driving. WorldDreamer builds image and video-based embedding through 3D convolution, joins text and action-based embedding, and accomplishes multimodal feature interaction through spatial-temporal patchwise self-attention and spatial-wise cross-attention. It realizes more types of video generation tasks and demonstrates a comprehensive understanding of visual dynamics in the general world.

### C. Diffusion-Based Methods

RaMViD [45] can perform video generation and infilling tasks by applying random masking of some frames in diffusion training. In Gen-1 [46], both the convolution block and the attention block incorporate a temporal layer to achieve explicit control of temporal consistency and to provide guidance for the diffusion process by extracting depth information and structural information.

In LidarDM [47], the driving scene guides the diffusion model to generate a 3D scene, which is combined with the dynamic agent to form a 4D world. Subsequent interaction between the agent and the 4D world generates realistic and temporally relevant point cloud videos. ADriver-I [48] introduces the concept of interleaved vision-action pair, matching unified vision signals and control signals, which are fed into multi-modal LLM and diffusion models, alternately generating new control signals and future scenario predictions. It realizes automated driving in a self-created world.

The data-driven approach in CTG [49] uses diffusion models to generate vehicle trajectories that are similar to human behavior. Explicit rules are introduced in the testing phase to enforce trajectories to remain realistic and physically feasible. The controllable simulation generated by the diffusion model based on real-world law can provide a more realistic simulation environment for self-driving vehicles. In CTG++ [50], LLM was added to enable the generation of complex traffic scenes controlled by simple language instructions. DrivingDiffusion [51] implements the generation of multi-view driving videos for a given 3D layout. The multi-view consistency of the generated video is ensured by information exchange between neighboring cameras. Cross-frame consistency is achieved by utilizing the first frame of a multi-view video sequence as a key control condition, and a local prompt is further used to guide the relationship between the global scene and local instances, so that instance generation quality can be improved. Panacea [52] uses a two-stage generation pipeline. In the first stage, realistic multi-view driving scene images are generated. In the second stage, the images are extended into video sequences. Panacea uses 4D attention to keep the scene consistent across time and views and adds bird's-eye view (BEV) features inserted by ControlNet to improve the controllability of video generation. Drive-WM [53] builds a world model that can generate multi-view videos with spatiotemporal consistency through the diffusion model and can implement end-to-end planning for autonomous driving. Its planning is based on the reward model's assessment of the future trajectories generated by the world model. WoVoGen [54] has two branches: the world model branch and the world volume-aware synthesis branch. The world model generates a world volume prediction based on past world volumes and the actions of the ego vehicle. World volume-aware can generate multi-camera video from the world volume through the diffusion model.

DriveDreamer [55] is a world model that can generate high-quality plausible driving scenarios, where diffusion models are used to build a comprehensive representation of the complex driving environment. Through a two-stage training process, the world model is equipped with an in-depth understanding of structured traffic constraints and the ability to predict future states. In DrivDreamer 2 [56], LLM is added, which allows users to describe the trajectory of agents in the driving scene through language and finally convert it into a high-quality driving scene video. Copilot4D [57] is a world model based on BEV tokens for discretizing diffusion processes. Copilot4D uses VQVAE to tokenize sensor observations and then predict the future through discrete diffusion, thereby solving the difficulties of complex and unstructured observation space and the need for generative models to decode multiple tokens in parallel.

UniSim [58] is an interactive generative world simulator built using the diffusion model. With multiple historical frames, the model can be instructed to generate new consecutive frames that match the intent of the interaction. UniSim enables the simulation of real-world experiences and can be used to simulate other artificial intelligence before they are generalized to the real world. SORA is a high-quality generative model for video. It maps video data to low-dimensional space by visual patch, extracts spacetime latent patches in video space, and uses a scalable transformer to learn the diffusion process based on the understanding of detailed text description of video. SORA can be regarded as a generalized simulator for the physical world.

### D. Other Methods

Exceptional approaches for building world models that have not been included before are listed in this chapter. TrafficGen [59] is a generative model that can generate realistic traffic scenarios. The model utilized a multi-context gating model as the decoder and trained individual multilayer perceptrons (MLPs) for placing the vehicle and forecasting their motion. Similarly, Tan et al. proposed the LCTGen [60] model, consisting of an LLM-Driven interpreter, a map retrieval model, and a generator. The interpreter model encoded the text description into a vectorized description of the scenarios, and the retrieval model searched for the corresponding map from the map dataset. In the generator, a query-based transformer model was designed to capture the interactions between agents and the map. Individual MLPs were trained for different prediction tasks, such as position prediction and motion prediction.

LeCun proposed the joint embedding predictive architectures (JEPA) [7] that was regarded as a better possible solution for implementing a general world model. Compared to other architectures, JEPA computes the loss of the samples in the latent space rather than the data space, which mitigates the model collapse. Based on the JEPA architecture, I-JEPA [61] and V-JEPA [62] were proposed to demonstrate the architecture's capability of pattern completion on image and video data.

### E. Evaluation Metrics

The selection of evaluation metrics for world models depends on the specific tasks and application domains in which the world models are employed. Existing evaluation metrics for assessing the performance of the world models can be divided into two approaches depending on the actual application: metrics that assess world models deployed within agents and those that evaluate the outputted data given by generative world models.

Metrics for evaluating world models deployed in agents are often equivalent to assessing the performance of agents that make decisions based on the world models' future predictions, typically reinforcement learning agents. In virtual gaming environments, such as Atari [63], Deepmind Control Suite [64], DMLab [65], and Minecraft. evaluations often utilize built-in environment scores or reward values. In autonomous driving scenarios, traffic violation and accident rates are two of the most common evaluation metrics.

The choice of metrics for evaluating generative world models is based on the modality of the data generated by the model. Common evaluation metrics for video-generating world models include Fréchet inception distance (FID) [66], Fréchet video distance (FVD) [67], inception score (IS) [68], and peak signal to noise ratio (PSNR) [69]. For models that generate driving traffic scenarios, the metrics include maximum mean discrepancy (MMD), mean average distance error (mADE),

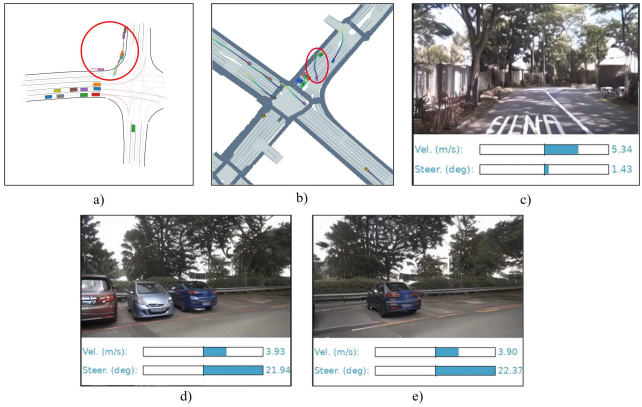
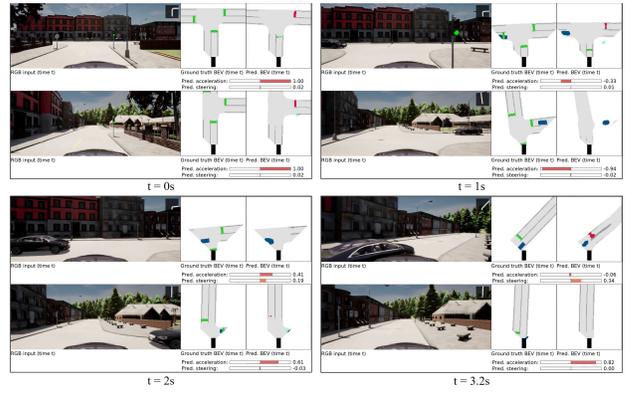

Fig. 2. Discovered unreasonable scenario imagination problems in existing world model for autonomous driving research. a) A traffic scenario generated by TrafficGen [59] where multiple vehicles are generated out of the drivable area. b) A traffic scenario generated by CTG [60] and the predicted participant behavior heavily violates the basic traffic rules. c) One single frame from a video scenario generated by DriveDreamer [55], where plausible but meaningless road marking is painted on the lane. d-e) Two subsequential frames in one video scenario generated by [55]. Some of the vehicles in the parking lot suddenly disappear from d) to e).

Fig. 3. A merging scenario in CARLA with two vehicles controlled by individual MILE [25] agents. The scenario starts at t = 0, where the merging vehicle (top) is trying to merge into the arriving vehicle's (down) driving lane. A collision occurs at t = 3.2s

mean final distance error (mFDE), and traffic rule violation metrics like scenario collision rate (SCR). In the case of 3D driving scene world models, which generate point clouds and 3D occupancy maps as future imagination, Chamfer distance, mean average precision (mAP), and intersection over union (IoU) related assessment functions are used as evaluation metrics.

## III. PATHOLOGY OF WORLD MODEL SAFETY

In this section, we discuss the inherent issues regarding world models for autonomous driving as a representative domain of embodied AI agents. We demonstrate the deficiencies through the concrete failing examples that were generated by selected models. We also provide a preliminary analysis of internal factors that may cause such failures. We concentrate on the scenario generation tasks as visualized by Fig. 2 in subsection A and explain our key findings on the failure of AI agents driven by world models in subsection B.

### A. Unreasonable Traffic Scenario Generated Through Selected World Models

Both TrafficGen [59] and CTG [60] are generative world models with the aim of generating realistic and naturalistic traffic scenarios. Fig. 2a) and b) show two traffic scenarios in BEV that are generated by TrafficGen and CTG, respectively. The red circle in Fig. 2a) indicates that part of the generated vehicles are not placed on the drivable regions initially, and the assigned trajectories are also out of the areas. In Fig. 2b), vehicles are driving in traffic lanes, but the driving behavior of the vehicle in the red circle heavily violates the traffic rules.

Besides the traffic scenarios in BEV, Fig. 2c) is a selected frame from a video scenario generated by DriverDreamer [55]. Even if the model performs well in generating a variety of realistic video scenarios, if one pays attention to important details such as road markings, one will still find that the video contains features that look plausible but are impossible to exist in the real world. Fig. 2d) and 2e) are two subsequent frames in one scenario. One can see that the red vehicle and silver vehicle in this parking lot disappear from d) to e). This kind of temporal inconsistency can also be found in most of the generative world models.

### B. Unsafe Behavior of AI Agents Driven by World Models

The world models also play a pivotal role in reinforcement learning or imitation learning training as the scene encoder and the future predictor. However, safety issues still exist when the world models misunderstand the behavior and intention of other interactive agents in rare or unseen scenarios. Fig. 3 demonstrates a merging scenario at a T-Junction in the CARLA simulator. The merging vehicle (top) and the arriving vehicle (down) are controlled by individual MILE agents, and the merging vehicle is trying to turn right to the arriving vehicle's future lane. The scenario is initialized at t = 0, and a crash between the merging vehicle and the arriving vehicle is identified at t = 3.2s. In this case, we adopted the testing methodology in [70] to find the critical configurations of the initial states.

## IV. RESEARCH DIRECTION PROPOSALS

We identified and discussed several highly prioritized research directions as a result of our investigation. It is clear that, as fundamental components of future safety-critical intelligent systems, current global models do not meet even the minimum safety requirements for autonomous agents. Guardrailing methods and controllable generative processes are indispensable for ultimate reliability and safety. Furthermore, we need trustworthy, technically feasible methods to quantitatively evaluate and benchmark the generative world models.

To be more specific, here is a non-exhaustive list of research challenges in data preparation, generative processes, and post-processing phases, respectively.

- **Quantitative Uncertainty Evaluations of Results:** Evaluating the confidence of outputs from world models through uncertainty measures is crucial for determining the reliability of generated results in downstream processes. Although uncertainty estimation in machine learning has been extensively studied, traditional methods such as Bayesian neural networks (BNNs) and conformal prediction may not be suitable due to the scale, parameter size, and computational overhead involved.
- **Symbolic Integration into Learning:** Incorporating prior knowledge to guide the generation process can be achieved through symbolic integration. Retrieval-augmented generation (RAG) [71], or, most recently, GraphRAG [72] can serve as a repository for embedded knowledge representations in vector form. Additionally, other neuro-symbolic approaches [73] can be employed to implement guardrails for the generative processes.
- **Controllable Learning Processes:** Utilizing hypernetworks [74] and even classical control theory [75] allow for the definition of predetermined learning goals. The learning processes can dynamically adapt if they deviate from these goals, ensuring that the predefined objectives are met consistently. There are abundant bodies of research on controllable machine learning which worthwhile to be explored [76].
- **Mechanistic Explainable Machine Learning:** Contrarily to classical explainability researches of machine learning, which focus on providing intuitive, high-level reasons and detailed mathematical formulations for a model's predictions, the mechanistic explainability [77] investigates internal behaviors of networks and try to understand their patterns, for example, activation of specific region of neurons could reveal whether a model is making particular decisions. Explainability is an effective auxiliary tool on the reliability issue. For instance, when the model hallucinates, a particular region of neurons could demonstrate more activities than others. Through such monitoring, countermeasures could be taken to ensure safety.
- **Benchmarks and Evaluation:** Establishing evaluation methods and benchmarks to measure the trustworthiness of generative results from world models [78]. According to various use cases of world models, multiple aspects must be considered. For example, if used for training dataset synthesis for autonomous driving systems, proper ways to measure the realism in terms of plausibility from both physical and social aspects need to be considered.

## V. Conclusion

World models have recently garnered significant attention within the AI research community as a foundational component for AI agents. In this paper, we present a comprehensive technical survey and analysis of the evolution of these technologies. Our results indicate that, despite the astonishing progress, the most recent world models still present numerous safety problems, making them unsuitable for safety-critical applications such as embodied artificial intelligence systems. Recent research from cognitive science reveals that language is more of communication than thoughts [79]. This could be a plausible explanation for why transformer- and autoregression-based language models alone would not be sufficient to develop world models that can truly capture the real world in its fullness, adequate research in engineering is therefore still required. Additionally, we identify and prioritize the key technical challenges that must be addressed to advance this field. In particular, while we recognize the potential of world models in the era of large language models (LLMs) for future intelligent systems driven by AI, substantial research efforts are still required to achieve this vision.